\documentclass[10pt, conference]{IEEEtran}

\usepackage[utf8]{inputenc} 
\usepackage[T1]{fontenc}    
\usepackage[hidelinks]{hyperref}       
\usepackage{url}            
\usepackage{booktabs}       
\usepackage{amsfonts}       
\usepackage{nicefrac}       
\usepackage{microtype}      
\usepackage{xcolor}         

\usepackage{amsmath,amssymb}
\usepackage{graphicx}
\usepackage{textcomp}
\usepackage{bm}
\usepackage[skip=1pt, belowskip=0pt]{caption}
\usepackage{subcaption}
\usepackage{cases}
\usepackage{cite}

\usepackage[ruled,vlined,linesnumbered]{algorithm2e} 
    \usepackage{setspace} 
\usepackage[makeroom]{cancel} 
\usepackage{amsthm}

\let\labelindent\relax
\usepackage{enumitem}  

\usepackage[normalem]{ulem} 

\def\BibTeX{{\rm B\kern-.05em{\sc i\kern-.025em b}\kern-.08em
    T\kern-.1667em\lower.7ex\hbox{E}\kern-.125emX}}
\AtBeginDocument{\definecolor{ojcolor}{cmyk}{0.93,0.59,0.15,0.02}}

\newcommand{\mytitle}{Regularization Trade-offs with Fake Features}

\title{\mytitle}

\newcommand{\Rbb}{\mathbb{R}}
\newcommand{\inrbb}[1]{\in\Rbb^{#1}}

\mathchardef\myhyphen="2D

\newcommand{\xbarhat}{\hat{\xbar}}
\newcommand{\xShat}{\hat{\xvec}_S}
\newcommand{\xChat}{\hat{\xvec}_C}
\newcommand{\xFhat}{\hat{\xvec}_F}

\newcommand{\Wbar}{\bar{\Wmat}}

\newcommand{\Amat}{\matr{A}}
\newcommand{\Atilde}{\Tilde{\matr{A}}}
\newcommand{\Abar}{\bar{\matr{A}}}
\newcommand{\AS}{\matr{A}_{S}}
\newcommand{\AC}{\matr{A}_{C}}
\newcommand{\AF}{\matr{A}_{F}}

\newcommand{\sigmav}{\sigma_{v}}

\newcommand{\omegaz}{\omega_{z}}

\newcommand{\yhat}{\hat{\yvec}}
\newcommand{\xhat}{\hat{\xvec}}

\newcommand{\xtilde}{\Tilde{\xvec}}

\newcommand{\xbar}{\bar{\xvec}}

\newcommand{\pbar}{\bar{p}}
\newcommand{\ptilde}{\tilde{p}}

\newcommand{\matr}[1]{\bm{#1}}

\newcommand{\Smat}{\matr{S}}
\newcommand{\Umat}{\matr{U}}
\newcommand{\Vmat}{\matr{V}}
\newcommand{\Wmat}{\matr{W}}

\newcommand{\avec}{\matr{a}}

\newcommand{\gvec}{\matr{g}}

\newcommand{\vvec}{\matr{v}}

\newcommand{\wvec}{\matr{w}}
\newcommand{\xvec}{\matr{x}}
\newcommand{\xS}{\matr{x}_S}
\newcommand{\xC}{\matr{x}_C}
\newcommand{\xF}{\matr{x}_F}
\newcommand{\yvec}{\matr{y}}
\newcommand{\zvec}{\matr{z}}

\newcommand{\Imat}{\matr{I}}

\newcommand{\eye}[1]{\Imat_{#1}}

\DeclareMathOperator{\Ebb}{\mathbb{E}}
\newcommand{\eunder}[1]{\underset{#1}{\Ebb}}

\newcommand{\Pbb}{\mathbb{P}}

\DeclareMathAlphabet{\mymathbb}{U}{BOONDOX-ds}{m}{n}

\newcommand{\diag}{\text{diag}}

\newcommand{\T}{^\mathrm{T}}
\newcommand{\p}{^+}

\newcommand{\inv}{^{-1}}

\let\oldin\in
\renewcommand{\in}{{\,\oldin\,}}

\let\oldnotin\notin
\renewcommand{\notin}{{\,\oldnotin\,}}

\newcommand{\irange}[2]{{i={#1},\,\dots,\,{#2}}}

\long\def\/*#1*/{}

\newcommand{\Nc }{\mathcal{N}}

\definecolor{myblue}{rgb}{0.3328, 0.3539, 0.7758}
\definecolor{mygreen2}{rgb}{ 0.0328 0.4758 0.0539} 
\definecolor{myred}{rgb}{0.85, 0.0328, 0.0539}

\newcommand{\rmax}{r_{\max}}
\newcommand{\rmin}{r_{\min}}

\newtheorem{thm}{\bf{Theorem}}
\newtheorem{cor}{\bf{Corollary}}

\newtheorem{rem}{\bf{Remark}}

\theoremstyle{remark}

\newenvironment{theorem}
{\par\noindent \thm \begin{itshape}\noindent}
{\end{itshape} \vspace{3pt}}

\newenvironment{corollary}
{\par\noindent  \cor \begin{itshape}\noindent}
{\end{itshape} \vspace{3pt}
}

\newenvironment{remark}
{\par\noindent \rem \begin{itshape}\noindent}
{\end{itshape} \vspace{3pt}}

\begin{document}

    \author{\IEEEauthorblockN{Martin Hellkvist, Ay\c ca \"Oz\c celikkale, Anders Ahl\'{e}n}
    \IEEEauthorblockA{\textit{Department of Electrical Engineering, }
    Uppsala University, Sweden \\
    \{Martin.Hellkvist, Ayca.Ozcelikkale, Anders.Ahlen\}@angstrom.uu.se}
    }

    \maketitle

    \begin{abstract}
        Recent successes of massively overparameterized models have inspired a new line of work investigating the underlying conditions that enable overparameterized models to generalize well.
        This paper considers a framework where the  possibly overparametrized model includes fake features,
        i.e., features that are present in the model but not in the data. 
        We present a non-asymptotic high-probability bound on the generalization error 
        of the ridge regression problem under the model misspecification of having fake features.
        Our high-probability results provide insights into the interplay between the implicit regularization provided by the fake features and the explicit regularization provided by the ridge parameter. 
        Numerical results illustrate the trade-off between the number of fake features and  how the optimal ridge parameter may heavily depend on the number of fake features. 
    \end{abstract}
    

    \section{Introduction}
Conventional wisdom in statistical learning suggests that the number of training samples should
exceed the number of model parameters in order to generalize well to data unseen during training.
However, 
it has recently been highlighted that the generalization error initially decreases with the model size in the underparametrized setting, 
and then again in the overparametrized setting,
hence the phenomenon of \textit{double descent} has been proposed \cite{belkin_reconciling_2019, belkin_two_2019}.

Double-descent behaviour can be caused by \textit{missing features}, 
i.e., features present in the data but not in the model \cite{belkin_two_2019, hastie_surprises_2020}.
Recently, 
surprising effects of additional irrelevant features in the model,
referred to as \textit{fake features},
i.e., features present in the model but not in the data, 
have been demonstrated \cite{chen2021multiple,hellkvist_2022_estimation_TSP, kobak_2020_optimal_ridge_can_be_negative, rao_misspecification_1971}.
In particular, 
 inclusion of fake features has been used to improve the estimation performance \cite{chen2021multiple, hellkvist_2022_estimation_TSP}.  
In this paper, we contribute to this line of work by providing high probability results in the finite regime for the generalization error associated with the ridge regression problem
and reveal insights into the trade-offs between the implicit regularization provided by the fake features and the explicit ridge regularization.

Observed for a wide range of models \cite{belkin_reconciling_2019},
the double-descent phenomenon in linear regression has been studied for the finite-dimensional case with Gaussian,
subgaussian and random features \cite{belkin_two_2019,Bartlett_benign_overfitting_2020,dascoli_2020_double_trouble, holzmuller_2021_on_the_universality,Liao_2021_a_random_matrix_analysis}
and in a Bayesian estimation setting \cite{hellkvist_2022_estimation_TSP},
as well as in the asymptotic high-dimensional setting \cite{hastie_surprises_2020}.
Extensions have been made to the investigation of optimal ridge parameter values \cite{kobak_2020_optimal_ridge_can_be_negative, nakkiran2020optimal},
and to the study of fake features \cite{chen2021multiple,hellkvist_2022_estimation_TSP, kobak_2020_optimal_ridge_can_be_negative}.
Trade-offs between explicit regularization 
and implicit regularization provided by different problem aspects have been investigated,
e.g., implicit regularization by asymptotic overparameterization \cite{kobak_2020_optimal_ridge_can_be_negative} 
and the equivalence of training noise and Tikhonov regularization in 
\cite{Bishop1995}.

Model misspecification often lead to double-descent curves \cite{belkin_two_2019, hastie_surprises_2020,hellkvist_2022_estimation_TSP}.
Robust methods under model misspecification have been focused in various works,
such as covariance matrix uncertainties in 
linear minimum mean-square error estimation \cite{lederman_tabrikian_constrained_2006, mittelman_miller_robust_2010},
and robust estimation with missing features \cite{Liu_Zachariah_Stoica_Robust_2020}. 
%

{\bf{Contributions:}} 
In this article, 
we contribute to the line of work with fake features under ridge-regression.
Our main contribution,
Theorem~\ref{thm:high-prob}, 
presents a high-probability bound for the generalization error of the finite-dimensional ridge regression problem with fake features. 
This is in contrast to earlier work which do not study regularization \cite{belkin_two_2019, Bartlett_benign_overfitting_2020}
or fake features \cite{bartlett_benign_ridge_2020},
study the asymptotic regime \cite{kobak_2020_optimal_ridge_can_be_negative, hastie_surprises_2020},
or provide results in terms of expectations over the regressor distribution \cite{hellkvist_2022_estimation_TSP}. 
Our result in Theorem~\ref{thm:high-prob} quantifies the trade-off between the fake features and the regularization parameter,
and provides insights into the mechanism behind this trade-off
through high-probability bounds on the eigenvalues.
Our focus on the i.i.d. Gaussian case allows us to provide clear expressions. 
Our numerical results quantify how 
the implicit regularization provided by the fake features
may compensate for a small ridge parameter in certain scenarios.

    \section{Problem Statement}\label{sec:problem_statement}
\subsection{Data generation:}
\kern-0.25em
The data comes from the following linear underlying system,
\begin{equation}\label{eqn:underlying_system}
    \yvec = \Atilde \xtilde + \vvec = \AS\xS + \AC\xC + \vvec,
\end{equation}
where $\yvec\!=\! [y_1,\cdots,y_n]\T\inrbb{n\times1}$
is the vector of outputs/observations,
$\xtilde\inrbb{\ptilde\times1}$ is the unknowns of interest
and 
$\vvec = [v_1,\cdots,v_n]\inrbb{n\times 1}$ is the vector of noise,
with 
$v_i\sim\Nc(0,\sigmav^2)$, $\forall i$, $\sigmav\geq 0$.
The feature matrix $\Atilde\inrbb{n\times \ptilde}$ 
is composed of the matrices
$\AS\inrbb{n\times p_S}$ and
$\AC\inrbb{n\times p_C}$,
as
\begin{equation}
\Atilde=[\AS,\AC],
\end{equation} 
with $\ptilde = p_S + p_C$.
The matrices $\AS$ and $\AC$ consist of identically and independently distributed (i.i.d.) standard Gaussian  entries $\sim \Nc(0,1)$, which are uncorrelated with the noise $\vvec$.
The vector of unknowns $\xtilde$ is composed of the components $\xS\inrbb{p_S\times 1}$ and $\xC\inrbb{p_C\times 1}$,
such that 
\begin{equation}
\xtilde = [\xS\T,\xC\T]\T.
\end{equation}
 
\subsection{Misspecified model:} While the data is generated by the underlying system in \eqref{eqn:underlying_system},
the estimation is performed based on the following misspecified model,
\begin{equation}\label{eqn:model_misspecified}
    \yvec = \Abar \xbar + \vvec = \AF\xF + \AS\xS + \vvec,
\end{equation}
where $\Abar\inrbb{n\times \pbar}$ is composed by 
\begin{equation}
\Abar = [\AF,\AS]
\end{equation}
with $\pbar = p_F + p_S$.
The matrix $\AF\inrbb{n\times p_F}$ has random i.i.d. standard Gaussian entries, statistically  independent of  $\AS$ and $\AC$.
The vector $\xbar$ is correspondingly composed as $\xbar = [\xF\T,\xS\T]\inrbb{\pbar \times 1}$,
where $\xF\inrbb{p_F\times 1}$.

We refer to the features in $\AF$, $\AS$ and $\AC$, as follows:
\begin{itemize}
    \item The features in $\AF$ are included in the misspecified model in \eqref{eqn:model_misspecified},
    but are irrelevant to the output variable $\yvec$,
    i.e., the data in \eqref{eqn:underlying_system},
    hence we refer to $\AF$ as \textbf{fake features}.
    
    \item The features $\AS$ are present both in the data generated by \eqref{eqn:underlying_system}
    and the misspecified model in \eqref{eqn:model_misspecified},
    hence we refer to them as \textbf{included underlying features}. 

    \item The features in $\AC$, 
    which are relevant to the data in $\yvec$, 
    are missing from the misspecified model in \eqref{eqn:model_misspecified}.
    Hence we refer to the features $\AC$ as \textbf{missing features}.
\end{itemize}

We employ the notation 
\begin{equation}
\Amat = [\AF, \AS, \AC]\inrbb{n\times p}
\end{equation}
to refer to the full set of features,
and correspondingly for the full set of unknowns, 
\begin{equation}
\xvec = [\xF\T,\xS\T,\xC\T]\T\inrbb{p \times 1},
\end{equation}
where $p=p_F+p_S+p_C$.

With the misspecified model in \eqref{eqn:model_misspecified}, 
we estimate $\xF$ and $\xS$ and 
we obtain the prediction of $\yvec$ as
\begin{equation}
\yhat = \AF\xFhat + \AS\xShat = \Abar\xbarhat \inrbb{n\times 1}.
\end{equation}

Recall that $\xbar = [\xF\T,\xS\T]$.
We obtain the estimate $\xbarhat$ by solving the following problem,
\begin{align}\label{eqn:ridgeregression}
    \xbarhat 
    & = \arg\min_{\xbarhat} \left\|\yvec - \left(\AF\xFhat + \AS\xShat\right)\right\|^2 + \lambda \|\xbarhat\|^2 \\
    & = \arg\min_{\xbarhat}  \left\|\yvec - \Abar\xbarhat\right\|^2 + \lambda\|\xbarhat\|^2,
\end{align}
where $\lambda \geq 0$ is the regularization parameter. 
Here, \eqref{eqn:ridgeregression} with $\lambda>0$ corresponds to the ridge regression problem whose solution is given by 
\begin{align}
    \xbarhat 
    =& \left( \Abar\T\Abar + \lambda \eye{p}\right)\inv \Abar\T \yvec 
    = \Abar\T\left(\Abar\Abar\T + \lambda \eye{n}\right)\inv \yvec.
    \label{eqn:ridge_estimate}
\end{align}
If $\lambda = 0$, 
we consider the minimum $\ell_2$-norm solution of \eqref{eqn:ridgeregression},
\begin{equation}\label{eqn:minnorm_estimate}
    \xbarhat = \Abar\p\yvec,
\end{equation}
where $(\cdot)\p$ denotes the Moore-Penrose pseudoinverse.
The estimate obtained by solving \eqref{eqn:ridgeregression} can be decomposed as 
\begin{equation}
     \xbarhat = \begin{bmatrix}
            \xFhat\T ~ \xShat\T
    \end{bmatrix}\T. 
\end{equation} 
Using $\xbarhat$, we obtain the estimate for $\xvec= [\xF\T,\xS\T,\xC\T]\T$ as follows,
\begin{equation}\label{eqn:xhat_final}
    \xhat = \begin{bmatrix}
            \xbarhat \\ \xChat
    \end{bmatrix} 
    = \begin{bmatrix}
             \xFhat \\ \xShat \\ \xChat
    \end{bmatrix} 
    =
    \begin{bmatrix}
            \xFhat \\ \xShat \\ \bm 0
    \end{bmatrix},
\end{equation}
where the estimate for the missing features is set to zero, 
i.e., $\xChat \! = \! \bm 0$,
as $\!\AC\!$ does not appear in the misspecified model \eqref{eqn:model_misspecified}. 

\subsection{Generalization Error:}
Suppose that we have obtained an estimate $\xhat$ as in \eqref{eqn:xhat_final}. A new unseen sample 
    $(y_*, \avec_*)$ comes
where $\avec_* = [\avec_{F*}\T, \avec_{S*}\T, \avec_{C*}\T]\T$. 
Hence,
    \begin{equation}
    y_* = \avec_{S*}\T\xS + \avec_{C*}\T\xC + v_* \inrbb{1\times 1},
    \end{equation}
where 
$\avec_{F*}\T\inrbb{1\times p_F}$, 
$\avec_{S*}\T\inrbb{1\times p_S}$,
and $\avec_{C*}\T\inrbb{1\times p_C}$ are
i.i.d. with the rows of $\AF$, $\AS$ and $\AC$ respectively,
and $v_*\inrbb{1\times 1}$ is i.i.d. with the noise samples in $\vvec$.
The corresponding prediction using $\xhat$ is 
 \begin{equation}
\hat{y}_* = \avec_{F*}\T\xFhat + \avec_{S*}\T\xShat.
\end{equation}
The \textit{generalization error} is given by
\begin{align}
    & J_y 
    = \eunder{y_*,\avec_*}\!\!\!\left[(y_* - \hat{y}_*)^2\right] \label{eqn:gen_y_def}  \\
    & = \!\! \eunder{y_*,\avec_*}\!\!\!\left[(\avec_{S*}\T\xS + \avec_{C*}\T\xC + v_* - \avec_{F*}\T\xFhat - \avec_{S*}\T\xShat)^2\right]\\
    &= \!\!\! \eunder{y_*,\avec_*}\!\!\!\left[\!\left(\![\avec_{F*}\T, \avec_{S*}\T, \avec_{C*}\T]\!\left(\!\begin{bmatrix} \bm 0 \\ \xS \\ \xC\end{bmatrix} \! - \!\begin{bmatrix} \xFhat \\ \xShat \\ \bm 0 \end{bmatrix}\right) \! + \! v_*\right)^2\right]\\
    &= \left\|
          \begin{bmatrix} \bm 0 \\ \xS \\ \xC\end{bmatrix} 
        - \begin{bmatrix} \xFhat \\ \xShat \\ \bm 0 \end{bmatrix}
    \right\|^2 + \sigmav^2  \label{eqn:gen_y_together}
\end{align}
We note that the generalization error 
consists of the respective errors in the components of $\xvec$ that correspond to the fake features $\AF$,
the included underlying features $\AS$ 
and the missing features $\AC$.

\begin{remark}\emph{(Interpolation with fake features)}
    \label{remark:interpolation}
    Recall that $\Abar = [\AF,\AS]\inrbb{n\times \pbar}$,
    and that $\pbar = p_F+p_S$,
    hence the estimate $\xbarhat = \Abar\p\yvec$ in \eqref{eqn:minnorm_estimate} is created using the fake features in $\AF\inrbb{n\times p_F}$ and included underlying features $\AS\inrbb{n\times p_S}$.
    If $n<\pbar$,
    then $\Abar\Abar\T$ is full rank with probability one (since entries of $\Abar$ are standard Gaussian i.i.d.),
    and the estimate $\yhat\inrbb{n\times 1}$ of the data $\yvec$
    is
    \begin{equation}
        \yhat = \AF\xFhat + \AS\xShat = \Abar\xbarhat = \Abar\Abar\p\yvec = \yvec,
    \end{equation}
    hence the training data is interpolated for $n < \pbar$,
    even when there are fake features in the misspecified model.
    Furthermore, we note that even if the misspecified model consists purely of fake features,
    i.e., if $p_S = 0$,
    and $n < p_F$,
    then we still have $\yhat = \yvec$.
    Hence, we still obtain interpolation without using any of the underlying features $\AS$ and $\AC$ in the estimation process. 
    We refer to the point where $n=\pbar$ as the \textit{interpolation threshold}.
    
\end{remark}

    \section{Generalization Error Bound}\label{sec:highProb}
In this section, 
we give our main result of the paper,
which is a high-probability bound on the generalization error $J_y$
in the finite-dimensional regime
for the ridge regression problem with $\lambda>0$. 
Note that here we analyze the generalization error $J_y$ in high probability with respect to training data  whereas $J_y$ itself is an average over test data.  

\begin{theorem}\label{thm:high-prob}
    Let the regularization parameter be nonzero, i.e., $\lambda > 0$,
    and 
    $t_1,t_2\geq 0$,
    $\rmax = \max(n,\pbar)$,
    $\rmin=\min(n,\pbar)$,
    and
    \begin{equation}\label{eqn:f_g-definition-main-result}
        f_g = \frac{(\sqrt{n}+\sqrt{\pbar}+t_2)^2}{((\sqrt{\rmax}-\sqrt{\rmin}-t_2)_+^2 + \lambda)^2},
    \end{equation}
    where $(\cdot)_+ = \max(\cdot,0)$,
    $(\cdot)_+^2 = ((\cdot)_+)^2$ and
    \begin{subnumcases}{ \Bar{f}_g = \label{eqn:bar_f_g_def_main-result} }
        \frac{\lambda^2}{((\sqrt{n}-\sqrt{\pbar}-t_2)_+^2 +\lambda)^2}, & if $n\geq \pbar$,\label{eqn:bar_f_g_def_main-result_n_geq_pbar}\\
        1, & if $n< \pbar$,
    \end{subnumcases}
    then the following holds for the generalization error in \eqref{eqn:gen_y_together},
    \begin{align}\label{eqn:high-prob-bound}
    \begin{split}
    & \Pbb\Big(
            J_y < \|\xS\|^2 \Bar{f}_g \\
            & \qquad\quad + \left(\|\xC\|^2+\sigmav^2\right) 
            f_g \left(\rmin + 2\sqrt{\rmin t_1} + 2 t_1 \right) \\
            & \qquad\quad + \left(\|\xC\|^2+\sigmav^2\right)
        \Big) > 1 - e^{-t_1} - 2e^{-t_2^2/2}.
    \end{split}
    \end{align}
    
\end{theorem}

\noindent Proof: See Appendix~\ref{proof:thm:high-prob}. %

    Note that if $t_2 \geq \sqrt{\rmax} - \sqrt{\rmin}$,
    then the denominators in \eqref{eqn:f_g-definition-main-result}
    and \eqref{eqn:bar_f_g_def_main-result_n_geq_pbar} reduces to $\lambda^2$.

    In Theorem~\ref{thm:high-prob}, both the upper bound on $J_y$ and the probability that the upper bound holds depend on $t_1$ and $t_2$. Hence, by varying $t_1$ and $t_2$, one obtains a series of upper bounds and associated probabilities.

From Theorem~\ref{thm:high-prob},
we observe the following:
\begin{enumerate}[align=left, style=nextline, leftmargin=*, topsep=1pt, itemsep=0pt, wide=\parindent]
    \item In order to avoid a very high value in the generalization error at the interpolation threshold $n=\pbar $ (Remark~\ref{remark:interpolation}),
    the ridge parameter $\lambda$ needs to be large enough.
    Otherwise,
    the probability parameter $t_2$ cannot be large enough to guarantee the bound in \eqref{eqn:high-prob-bound} holds
    without making the bound very large due to the denominators being too small in \eqref{eqn:f_g-definition-main-result} and \eqref{eqn:bar_f_g_def_main-result_n_geq_pbar}. %

    \item In addition to the explicit ridge regularization,
    the fake features in $\AF$ have a regularizing effect on the error bound.
    Suppose that $n\approx p_S$ and $\lambda$ is very small,
    hence the problem without fake features is close to the interpolation threshold at $n=p_S$,
    and the bound in \eqref{eqn:high-prob-bound} is very large.
    If there are enough fake features,
    then the actual problem dimensions will be far away from the threshold $n\approx \pbar$,
    hence the bound will take on smaller values.
    Nevertheless,
    if the regularization parameter $\lambda$ is large enough,
    then the bound takes on small values regardless of the presence of fake features.
\end{enumerate}

\begin{remark}
    Theorem~\ref{thm:high-prob} quantifies the trade-off  between the number of fake features and the ridge parameter using a high-probability bound. 
    In contrast to the works that study the regression problem without regularization 
    \cite{belkin_two_2019, Bartlett_benign_overfitting_2020} 
    or regularization in the asymptotic high-dimensional regime 
    \cite{kobak_2020_optimal_ridge_can_be_negative, hastie_surprises_2020},
    in  terms of expectation over the regressor distribution
    \cite{hellkvist_2022_estimation_TSP},
    here we provide high-probability bounds that consider the presence of both fake features and regularization.

\end{remark}

%

\section{Numerical Results}

\subsection{Details of the Numerical Simulations}\label{sec:appendix:numerical}
    In the following simulations
    we compute empirical averages for the generalization error.
    We now describe how we obtain these averages for a given set of problem dimensions $n$, $p_F$, $p_S$ and $p_C$,
    the fixed power ratio coefficient $r_S$,
    and the noise level $\sigmav^2$,
    and total power $P$ of the underlying unknowns.

    We generate the underlying unknowns $\xS$ and $\xC$ as 
    $ 
        \xS = \sqrt{r_S \tfrac{P}{p_S} } \bm 1 \inrbb{p_S\times 1},~
        \xC = \sqrt{(1-r_S)\tfrac{P}{p_C}} \bm 1\inrbb{p_C\times 1},
    $ 
    where $\bm 1$ denotes a vector of ones with appropriate dimensions. 
    For the test data, we have $n_{test}=20000$ samples.
    We generate $M=100$ realizations of the training feature matrices
    $\AF$, $\AS$, $\AC$,
    as well as corresponding test feature matrices $\Amat_{F,test}\inrbb{n_{test}\times p_F }$, $\Amat_{S,test}\inrbb{n_{test}\times p_S}$ and $\Amat_{C,test}\inrbb{n_{test}\times p_C}$.
    The feature matrices are all i.i.d. standard Gaussian matrices.
    For each of these $M$ sets we generate $M$ noise vectors $\vvec\inrbb{n\times 1}$ and $\vvec_{test}\inrbb{n_{test}\times 1}$,
    with standard Gaussian entries, scaled with $\sigmav$.
    We generate the corresponding training and test data as
    $
        \yvec = \AS\xS + \AC\xC + \vvec, ~
        \yvec_{test} = \Amat_{S,test} \xS + \Amat_{C,test}\xC + \vvec_{test}.
    $
    We then compute $\xbarhat$ as the solution to \eqref{eqn:ridgeregression}, 
    i.e.,
    $
        \xbarhat = \Abar\T(\Abar \Abar\T + \lambda\eye{n})\p\yvec.
    $
    which corresponds to the minimum-norm solution for $\lambda=0$. 
    The predictions of the test data is computed as
    $
        \yhat_{test} = \Amat_{F,test}\xFhat + \Amat_{S,test}\xShat,
    $
    and the corresponding error instance as 
    $
        J_{y} = \|\yvec_{test} - \yhat_{test}\|^2 - \sigma_v^2,
    $
    which is then averaged over the $M$ sets of noise vectors,
    and then as well over the $M$ sets of feature matrices.
    We have $n=200$,
    and the number of included and missing features is $p_S = p_C = 100$, $\sigmav=10$,
    the signal power in $\xtilde$ is $\|\xtilde\|^2=200$,
    and ratio of the power in the included underlying unknowns $\xS$ is $\frac{\|\xS\|^2}{\|\xtilde\|^2} = r_S = 0.5$.

\begin{figure}
    \centering
    \includegraphics[width=0.9\linewidth]{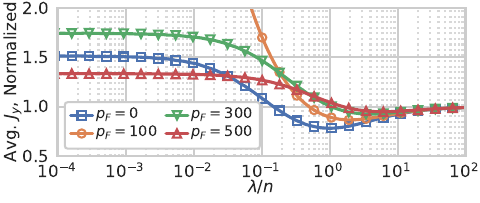}
    \caption{
        The empirical average of the generalization error $J_y$ 
        versus the ridge parameter $\lambda$.
        }
    \label{fig:5f_y}
\end{figure}    
\begin{figure}
    \centering
    \includegraphics[width=0.9\linewidth]{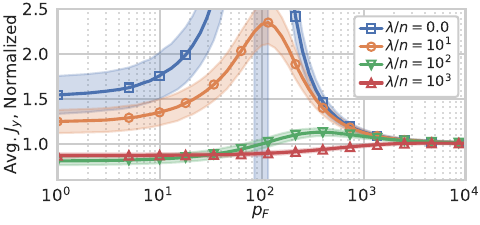}
    \caption{The empirical average of the generalization error $J_y$ (solid lines) $+/-$ one standard deviation (shaded areas) versus the number of fake features $p_F$. 
        }
    \label{fig:6b}
\end{figure}

\subsection{Trade-offs between the regularization parameter $\lambda$ and the number of fake features}
We investigate the effect that ridge regularization has on the problem under the presence of the fake features in $\AF$ 
by plotting the average generalization error $J_y$ in Figure~\ref{fig:5f_y} and \ref{fig:6b},
obtained via simulation of the problem in \eqref{eqn:minnorm_estimate}.
In Figure~\ref{fig:5f_y},
we plot the empirical average generalization error versus the ridge parameter 
$\lambda$,
for varying number of fake features $p_F$.
In Figure~\ref{fig:6b} we plot the error versus $p_F$,
for varying values of $\lambda$.
The shaded areas in Figure~\ref{fig:6b} indicate the standard deviations.

These figures support the following conclusions:
    \textit{i)} It is possible to decrease the error by increasing the number of fake features. 
    \textit{ii)} The best choice of $\lambda$ depends on the number of fake features in the model.
    The effect of fake features can be interpreted as the fake features providing implicit regularization to the problem;
    and that the regularization provided by fake features can be used to compensate for low levels of $\lambda$,
    i.e., low levels of explicit regularization, for some scenarios. 
    
For \textit{i)},
see  Figure~\ref{fig:6b}  for small values of $\lambda$, 
i.e., $\lambda/n \in \{0, 10\}$: Here the lowest error over all $p_F$ is achieved by increasing $p_F$ to $p_F>10^3$,
rather than having $p_F$ small.
Hence,  having a large number of fake features $p_F$ can compensate for having a small ridge parameter $\lambda$ by providing implicit regularization to the problem.
For \textit{ii)},
the plot in Figure~\ref{fig:5f_y} illustrates that the best choices of $\lambda$,
i.e., the locations of the local minima of the respective curves,
increase as the number of fake features $p_F$ is increased,
as well the values of these minima.
Hence, if the explicit regularization parameter $\lambda$ is large enough,
then the problem without fake features has enough regularization,
and the smallest possible number of fake features $p_F$ gives the lowest error.
These observations illustrate that a large number of fake features may compensate for having too little explicit regularization,
but if there is enough explicit regularization,
then a higher number of fake features may increase the error.

As shown in Theorem~\ref{thm:high-prob},
$\lambda$ should be large enough in order to bound the generalization error $J_y$ around the interpolation threshold with high probability.
We observe this effect in Figure~\ref{fig:6b},
where the large enough values of $\lambda$,
i.e., $\lambda/n\in\{10^2, 10^3\}$,
dampen the peak in error around the interpolation threshold,
that is otherwise seen for the smaller values of $\lambda$.
Furthermore,
we note that if $\lambda=0$,
then the standard deviation is extremely large around the interpolation threshold of $p_F=100$,
and if $\lambda$ increases,
then the standard deviation decreases.
In general, higher values of $\lambda$ decrease the standard deviation, 
i.e., the variation around the mean value. 
Similarly, increasing $p_F$ decreases the variance, 
e.g., compare $\lambda/n =0$ curve for $p_F \approx 0$ and $\approx 10^3$.
This again suggests that $p_F$ can have a regularizing effect,
similar to the ridge parameter $\lambda$.   

\section{Conclusions}\label{sec:conclusions}

We provide a non-asymptotic high-probability bound for the generalization error of the ridge regression solution when an arbitrary number of fake features are present.
This result reveals analytical insights on the interplay between the implicit regularization provided by the fake features and the explicit regularization provided by the ridge regularization. 

We have considered  linear models with  isotropic Gaussian features. Extensions  into non-linear models with more general feature covariance structures and other regularization frameworks are considered important research directions.

\setlength{\abovedisplayskip}{3pt}
\setlength{\belowdisplayskip}{3pt}
\appendix
\section{Appendix}

\subsection{Proof of Theorem~\ref{thm:high-prob}}
\label{proof:thm:high-prob}
    With $\xbarhat = \Wbar \yvec$,
    we denote the estimator as
    \begin{equation}\label{eqn:wbar_ridge_proof}
        \Wbar = \Abar\T(\Abar\Abar\T + \lambda\eye{n})\inv.
    \end{equation}
    We denote the full singular value decomposition of $\Abar$ by
    \begin{equation}\label{eqn:abar_svd_proof}
        \Abar = \Umat \Smat \Vmat\T,
    \end{equation}
    where $\Umat \inrbb{n\times n} $ and $\Vmat\inrbb{\pbar \times\pbar}$ are orthogonal matrices,
    and the diagonal matrix $\Smat\inrbb{n\times \pbar}$ contains the singular values $s_i$ of $\Abar$,
    $\irange{1}{\min(n,\pbar)}$.
    We let $s_i=0$ if $i>\min(n,\pbar)$.

    From \eqref{eqn:gen_y_together}, 
    we have
   \begin{align}
        & J_y
        = \left\| \begin{bmatrix} \bm 0 \\ \xS \end{bmatrix}
        - \begin{bmatrix} \xFhat \\ \xShat \end{bmatrix}\right\|^2
        + \|\xC\|^2 + \sigmav^2 \\
        & = \! \left\| \! \begin{bmatrix} \bm 0 \\ \xS \end{bmatrix} \! 
        - \! \Wbar\left( \AS\xS + \AC\xC + \vvec \right)\right\|^2
        \! \! \! + \! \|\xC\|^2 \! + \! \sigmav^2 \\
        & = \left\| \left(\eye{\pbar} - \Wbar\Abar\right)\begin{bmatrix} \bm 0 \\ \xS \end{bmatrix}
        - \Wbar\zvec\right\|^2
        + \omegaz^2 .
    \end{align}
    Here we introduced the vector $\zvec = \AC\xC + \vvec \T\inrbb{n\times 1}$,
    with the entries $[z_1,\cdots,z_n]$
    which are i.i.d. random variables with $\Nc(0,\omegaz^2)$,
    and $\omegaz^2 = \|\xC\|^2 + \sigmav^2$.

    Using the triangle inequality 
    (for two vectors $\vvec$, $\wvec$,
    $\|\vvec - \wvec\|^2 \leq 2\|\vvec\|^2+2\|\wvec\|^2$),
    as well as the submultiplicativity of the $\ell_2$-norm,
    we have
        $ J_y \leq 2\left\|\eye{\pbar} - \Wbar\Abar\right\|^2\|\xS\|^2
        + 2\left\| \Wbar\zvec\right\|^2
        + \omegaz^2. $
    We continue by plugging in \eqref{eqn:wbar_ridge_proof} and \eqref{eqn:abar_svd_proof},
    \begin{align}
    \begin{split}
        & J_y \leq 2\left\|\eye{\pbar} - \Smat\T(\Smat\Smat\T\! +\! \lambda\eye{n})\inv\Smat\right\|^2\|\xS\|^2 \\
        & \quad + 2\left\| \Smat\T(\Smat\Smat\T\!+\!\lambda\eye{n})\inv\Vmat\T\zvec\right\|^2
        \!\!+\! \omegaz^2 \\
    \end{split} \\
    \begin{split}
        & \sim 2\left\|\eye{\pbar} - \Smat\T(\Smat\Smat\T\! +\! \lambda\eye{n})\inv\Smat\right\|^2\|\xS\|^2 \\
        & \quad + 2\left\| \Smat\T(\Smat\Smat\T\!+\!\lambda\eye{n})\inv\zvec\right\|^2
        \!\!+\! \omegaz^2
    \end{split}
    \end{align}    
    where we used the unitary invariance of the norm,
    and $\Vmat\T\zvec \sim \zvec$ due to the rotational invariance of the distribution of $\zvec$.
    
    We continue by utilizing the diagonal structure of $\Smat$,
    \begin{align}
    \begin{split}
        J_y 
        & \leq 2\left\|\eye{\pbar}-\diag\left(\tfrac{s_i^2}{s_i^2 + \lambda}\right)\right\|^2\|\xS\|^2 \\
        & \quad + 2\left\| \left[\tfrac{s_1}{s_1^2 + \lambda}z_1, \,\cdots, \,\tfrac{s_{\rmin}}{s_{\rmin}^2 + \lambda}z_{\rmin}\right]\T\right\|^2
        \!\!+\! \omegaz^2 
    \end{split} \\
        & = 2\left\|\diag\left(\tfrac{\lambda^2}{(s_i^2 + \lambda)^2}\right)\right\|\|\xS\|^2
        + 2 \sum_{i=1}^{\rmin} g_i \, z_i^2
        + \omegaz^2 \label{eqn:proof:J_y_bound_sum_gi_zi^2}
    \end{align}
    where 
    $\irange{1}{\pbar}$ in the first term
    and $s_i = 0$ if $i>\rmin\triangleq \min(n,\pbar)$,
    and where we have introduced the coefficients
    \begin{equation}\label{eqn:proof:gi_def}
        g_i = \tfrac{s_i^2}{(s_i^2+\lambda)^2}, ~\irange{1}{\rmin}.
    \end{equation}
    
    We now focus on the second term of \eqref{eqn:proof:J_y_bound_sum_gi_zi^2}.
    The following corollary can be derived from \cite[Lemma 1]{Laurent2000AdaptiveEO}:
    \begin{corollary}\label{col:chi-squared-summation}
        Let $z_i$, $\irange{1}{r}$, be i.i.d. with $z_i \sim \Nc(0,\omegaz^2)$, 
        and let $\gvec = [g_1, \cdots, g_r]\T\inrbb{r\times 1}$,
        with $g_i>0$, $\forall i$,
        and $t>0$.
        Consider the event
        $
            E \! = \! \left\{\sum_{i=1}^r g_i z_i^2 
                < \omegaz^2\left(\sum_{i=1}^{r} g_i + 2\|\gvec\|\sqrt{t} + 2\|\gvec\|_{\infty} t \right)\right\},
        $
        where $\|\gvec\|_{\infty} = \sup_{\irange{1}{r}} g_i$.
        Then,
        $
            \Pbb\left(E \right) 
            \geq 1 - e^{-t}.
        $
        
    \end{corollary}
    With $t_1>0$
    and $g_i$ and $z_i$ as in \eqref{eqn:proof:J_y_bound_sum_gi_zi^2},
    we denote the event
    \begin{equation}\label{eqn:proof:E_1_def}
        \!\!\!\!\!
        E_1 \!
        = \! \left\{\!
            \sum_{i=1}^{\rmin} g_i z_i^2 \!\!
            < \! \omegaz^2 \! 
            \left(
                \sum_{i=1}^{\rmin} g_i \!
                + \! 2\|\gvec\|\sqrt{t_1} + 2\|\gvec\|_{\infty} t_1\!
            \right)\!\!
        \right\},
    \end{equation}
    where $\gvec =[g_1,\cdots,g_{\rmin}]\T\inrbb{\rmin\times 1}$,
    and from Corollary~\ref{col:chi-squared-summation},
    we have that
    \begin{equation}\label{eqn:proof:E_1-probability}
        \Pbb\left( E_1 \right) 
        > 1 - e^{-t_1}.
    \end{equation}
    We note that the variables $g_i$ in \eqref{eqn:proof:gi_def} are random over the singular values $s_i$ of $\Abar\inrbb{n\times \pbar}$,
    and we continue by upper bound these $g_i$ with a high-probability bound based on the distribution of $s_i$.
    We begin by noting that 
    for each $g_i$,
    \begin{equation}\label{eqn:proof:g_i_upper_bound_smax_smin}
        g_i \leq \tfrac{s_{\max}^2}{(s_{\min}^2 + \lambda)^2},
        \irange{1}{\rmin}.
    \end{equation}
   We denote the event $E_{2a}$ to bound the singular values as
    \begin{align}\label{eqn:proof:E2'_def}
        \!\!\!\!\! E_{2a} \!\! = \!\! \big\{\!
            \sqrt{\rmax} \! - \!\! \sqrt{\rmin}\! -\! t_2 \!
            \leq \! s_{\min} \!
            \leq \! s_{\max} \! \leq \! \sqrt{n}\!
            + \! \sqrt{\pbar} \! + \! t_2 \!
        \big\}\! ,
    \end{align}
    where $s_{\min}$ and $s_{\max}$ denotes the smallest and the largest singular values of $\Abar$, respectively,
    and $\rmax = \max(n,\pbar)$,
    and $\rmin = \min(n,\pbar)$,
    as defined previously.
    Using \cite[eqn. (2.3)]{rudelson_non-asymptotic_2010},
    we have that for any $t_2\geq 0$,
    \begin{equation}\label{eqn:tracy-widom}
        \Pbb\left(
            E_{2a}
        \right)
        \geq 1 - 2 e^{-t_2^2/2}.
    \end{equation}
    We will use this probability bound later in the proof to find the desired probability bound on $J_y$.
    
    We now define $f_g$ by plugging in the lower and upper bounds of \eqref{eqn:proof:E2'_def} into the bound in \eqref{eqn:proof:g_i_upper_bound_smax_smin},
    \begin{equation}\label{eqn:f_g-definition}
        f_g = \tfrac{(\sqrt{n}+\sqrt{\pbar}+t_2)^2}{((\sqrt{\rmax}-\sqrt{\rmin}-t_2)^2 + \lambda)^2}.
    \end{equation}
    We now define the event $E_2$ using \eqref{eqn:proof:g_i_upper_bound_smax_smin} and 
    \eqref{eqn:f_g-definition},
    \begin{equation}\label{eqn:proof:E2_def}
        E_2 = \left\{ g_i \leq f_g \right\},
    \end{equation}
    where $E_{2a}\Rightarrow E_2$.
    We combining the events $E_1$ in \eqref{eqn:proof:E_1_def} and $E_2$ in \eqref{eqn:proof:E2_def},
    to obtain the event $E_3$ as
    \begin{equation}\label{eqn:proof:E_3_def}
        \! E_3 = \! \left\{
            \sum_{i=1}^{\rmin} g_i z_i^2 < \omegaz^2 f_g \left( \rmin + 2 \sqrt{\rmin t_1} + 2 t_1 \right)
        \right\},
    \end{equation}
    where $E_1\cap E_2 \Rightarrow E_3$.
    %
    We now continue with the leading term of \eqref{eqn:proof:J_y_bound_sum_gi_zi^2},
    which is bounded as
    $
       \left\|\diag\left(\frac{\lambda^2}{(s_i^2+\lambda)^2}\right)\right\|
        \leq \frac{\lambda^2}{(s_{\min}^2 + \lambda)^2}.
    $
    where $\irange{1}{\pbar}$.
    We recall that if $i>\min(n,\pbar)$ then $s_i = 0$.
    Hence if $n < \pbar$, we define
    $E_4$ as follows
    \begin{equation} \label{eqn:xs_component_n_leq_pbar_hp_proof}
        E_4 = \left\{\left\|\diag\left(\tfrac{\lambda^2}{(s_i^2+\lambda)^2}\right)\right\| = 1\right\}.
    \end{equation}
    If instead $n\geq \pbar$,
    then we define
    \begin{equation}\label{eqn:xs_component_n_geq_pbar_hp_proof}
     \!\!\!\!\!  
        E_4 \! = \! \left\{ \!
            \left\| \diag\!\left(\!\tfrac{\lambda^2}{(s_i^2+\lambda)^2}\!\right)\!\right\| \!
            \leq \! \tfrac{\lambda^2}{((\sqrt{n} \! - \! \sqrt{\pbar} \! - \! t_2)^2 \! + \! \lambda)^2}
        \right\}
    \end{equation}
    and note that $E_{2a}\Rightarrow E_4$.
    We combine \eqref{eqn:xs_component_n_leq_pbar_hp_proof},
    \eqref{eqn:xs_component_n_geq_pbar_hp_proof}
    and \eqref{eqn:proof:E_3_def}
    with the bound on $J_y$ in \eqref{eqn:proof:J_y_bound_sum_gi_zi^2} to obtain 
    $ E_5 = \Big\{
        J_y < \|\xS\|^2 \Bar{f}_g
        + \left(\|\xC\|^2+\sigmav^2\right) 
            \big(f_g \left(\rmin \! + \! 2\sqrt{\rmin t_1} \! + \! 2 t_1 \right) + 1\big)
        \Big\}.
        $
    with $t_1,t_2\geq 0$, 
    $f_g$ as in \eqref{eqn:f_g-definition},
    and where
    \begin{subnumcases}{\bar{f}_g = }
        \tfrac{\lambda^2}{((\sqrt{n}-\sqrt{\pbar}-t_2)^2 +\lambda)^2} &  if $n\geq \pbar$,\\
        1 & if $n< \pbar$.
    \end{subnumcases}

    We note that
    i) $E_1$ is independent from $E_2$ and $E_4$,
    ii) $E_{2a}\Rightarrow E_2$ and $E_{2a}\Rightarrow E_4$,
    hence $E_{2a}\Rightarrow E_2 \cap E_4$,
    and if we denote $E_{24} = E_2 \cap E_4$,
    then by \eqref{eqn:tracy-widom} we can write
    \begin{equation}
        \Pbb(E_{24}) \geq \Pbb(E_{2a}) \geq 1 - 2e^{-t_2^2/2},\quad \Pbb(E_{24}^c) \leq 2e^{-t_2^2/2}.
    \end{equation}  
    By \eqref{eqn:proof:E_1-probability} we have that
    $
        \Pbb(E_1^c) \leq e^{-t_1}.
    $
    Furthermore, we have
    \begin{align}
        \Pbb(E_5) 
        & \geq \Pbb(E_3 \cap E_4)
        \geq \Pbb(E_1 \cap E_2 \cap E_4) \\
        & = \Pbb(E_1 \cap E_{24})
        = 1 - \Pbb(E_1^c \cup E_{24}^c) \\
        & \geq 1 - \Pbb(E_1^c) - \Pbb(E_{24}^c)
        \geq 1 - e^{-t_1} - 2e^{-t_2^2/2},
    \end{align}
    where we have used the union bound to obtain
    $\Pbb(E_1^c \cup E_{24}^c) \leq \Pbb(E_1^c) + \Pbb(E_{24}^c)\leq e^{-t_1} + 2e^{-t_2^2/2}$.
    This concludes the proof.

    {
        \bibliographystyle{IEEEtran}
        \bibliography{ref,ref2}   
    }

\end{document}